\documentclass[a4paper]{llncs}
\usepackage{multirow}
\usepackage{listings}
\usepackage{float}
\usepackage{array}
\usepackage{color}
\usepackage{amssymb} 
\usepackage{amsmath}
\usepackage{paralist}
\usepackage{booktabs} 
\usepackage{times}
\usepackage{graphicx}
\usepackage{verbatim}
\usepackage{epsfig}
\usepackage{url}
\usepackage{color}
\usepackage{alltt}
\usepackage{latexsym}
\usepackage{cite}
\usepackage{listings}
\usepackage[dvipsnames]{xcolor}

\newcommand{\Caption}[1]{\vspace*{-1em}\begin{footnotesize}\caption{#1}\end{footnotesize}\vspace*{-3ex}}

\newcommand{\Comment}[1]{}

\newcommand{\Section}[1]{\vspace*{-1ex}\section{#1}\vspace*{-1ex}}

\newenvironment{CodeOut}{\vspace*{0em}\begin{scriptsize}}
                        {\end{scriptsize}\vspace*{0em}}

\usepackage{xspace}
\newcommand{\score}{\text{score}\xspace{}}
\newcommand{\cen}{\text{cen}\xspace{}}

\newtheorem{hypothesis}{Hypothesis}

\usepackage{tikz,ifthen,pgfplots}
\usetikzlibrary{arrows,trees,backgrounds,automata,shapes,decorations,plotmarks,fit,calc,positioning,shadows,chains}
\tikzstyle{every pin edge}=[<-,shorten <=1pt]
\tikzstyle{neuron}=[circle,fill=black!25,minimum size=17pt,inner sep=0pt]
\tikzstyle{input neuron}=[neuron, fill=green!50]
\tikzstyle{output neuron}=[neuron, fill=red!50]
\tikzstyle{hidden neuron}=[neuron, fill=blue!50]
\tikzstyle{annot} = [text width=4em, text centered]
\usetikzlibrary{calc}

\usepackage{aircraftshapes}
\usetikzlibrary{angles,quotes}
\usepgfplotslibrary{groupplots}

\begin{document}

    \title{DeepSafe: A Data-driven Approach for Checking Adversarial Robustness in Neural Networks}
\author{Divya Gopinath\inst{1}  \and
        Guy Katz\inst{2} \and
        Corina S. P\u{a}s\u{a}reanu\inst{1} \and
        Clark Barrett\inst{2}}

\institute{
Carnegie Mellon University Silicon Valley  \\
\email{divgml@gmail.com,\\corina.pasareanu@west.cmu.edu, corina.s.pasareanu@nasa.gov}
\and
Stanford University \\
\email{guyk@cs.stanford.edu, barrett@cs.stanford.edu}
}

\maketitle

    \begin{abstract}
Deep neural networks have become widely used, obtaining remarkable results in
domains such as computer vision, speech recognition, natural language
processing, audio recognition, social network filtering, machine translation,
and bio-informatics, where they have produced results comparable to human
experts. However, these networks can be easily ``fooled'' by \emph{adversarial
perturbations}: minimal changes to correctly-classified inputs, that cause
the network to misclassify them. This phenomenon represents a concern for both
safety and security, but it is currently unclear how to measure a network's
robustness against such perturbations. Existing techniques are limited to checking robustness
around a few individual input points, providing only very limited
guarantees.  We propose a novel approach for automatically identifying
{\em safe regions} of the input space, within which the network is robust against
adversarial perturbations. The approach is data-guided, relying on
clustering to identify well-defined geometric regions as candidate safe regions.
We then utilize verification
techniques to confirm that these regions are safe or to provide
counter-examples showing that they are not safe.
We also introduce the notion of
\emph{targeted robustness} which, for a given target label and region, ensures
that a NN does not map any input in the region to the target label.
\Comment{ to be examined by a human or to be used in retraining
  of the network}We evaluated our technique on the MNIST dataset and on a
neural network implementation of a controller for the next-generation Airborne
Collision Avoidance System for unmanned aircraft (ACAS Xu). For these networks,
our approach identified multiple regions which were completely safe as well as
some which were only safe for specific labels.  It also discovered several
adversarial perturbations of interest.
\end{abstract} 

   \Section{Introduction}
\label{sec:intro}

In recent years, advances in deep neural networks (NN) have enabled the
representation and modeling of complex non-linear
relationships. In this paper, we study a common use of NN as classifiers that take in complex, high dimensional input, pass
it through multiple layers of transformations, and finally assign to it a specific output label or class.
Such classifiers have been used in a variety of applications, including pattern analysis, image
classification, speech and audio recognition, and self-driving cars; it is
expected that this trend
will continue and intensify, with neural networks also being integrated into
safety-critical systems which require high assurance guarantees.

While the usefulness of neural networks is evident, it
has been observed that state-of-the-art networks are highly vulnerable
to \emph{adversarial perturbations}: given a correctly-classified input $x$, it is
possible to find a new input $x'$ that is very similar to $x$ but is
assigned a different label~\cite{SzegedyZSBEGF13}. For instance, in image-recognition networks it is possible to
add a small amount of noise (undetectable by the human eye) to an
image and change how it is classified by the network.

Worse still, adversarial examples have also been
found to \emph{transfer} across networks, making it possible to attack
networks in a black-box fashion, without access to their
weights. Recent work has demonstrated that such attacks can be
carried out in practice~\cite{KuGoBe16}. Vulnerability of neural networks to
adversarial perturbations is thus a safety and security concern, and it is
essential to explore systematic methods for evaluating and improving
the robustness of neural networks against such attacks.

To date, researchers have mostly focused on efficiently
finding adversarial perturbations around select individual input points.
The problem is typically cast as an optimization problem: for a given network $F$
and an input $x$, find an $x'$ for which $F(x')\neq F(x)$ while
minimizing $\lVert x - x' \rVert$. In other words, the goal is to find an input
$x'$ as close as possible to $x$ such that $x'$ and $x$ are labeled
differently. Finding the optimal solution for this optimization problem is
computationally difficult, and so various approximation approaches
have been proposed. Some approaches are \emph{gradient based}~\cite{SzegedyZSBEGF13,GoodfellowSS14,FeinmanCSG17},
whereas others use optimization techniques~\cite{Carlini017}.
There are also techniques that focus on generating \emph{targeted attacks}:
adversarial perturbations that result in the network classifying the perturbed
input with a specific target label~\cite{SzegedyZSBEGF13,GoodfellowSS14,FeinmanCSG17}.

These various approaches for finding adversarial perturbations have successfully demonstrated the weakness of many
state-of-the-art networks; however, because these approaches operate on individual
input points, it is unclear how to apply them to large input domains, unless
one does a brute-force enumeration of all input values which is infeasible for
most input domains.
Furthermore, because they are inherently incomplete, these techniques do not
provide any robustness guarantees when they fail to find an adversarial input.
Orthogonal approaches have also been proposed for training networks that
are robust against adversarial perturbations, but these, too, provide no
formal assurances~\cite{PapernotM16}.

Formal methods provide a promising way for providing such guarantees.
Recent approaches tackle neural network verification~\cite{HuangKWW17,KaBaDiJuKo17Reluplex}
by casting it as an SMT solving problem. 
Although typically slower than the aforementioned techniques,
verification can provide sound assurances that no adversarial examples exist
within a given input domain. Still,
these techniques do
not provide any guidance on how to select meaningful regions within which the network is expected to behave consistently.
 And although it is possible to formulate a naive notion of {\em global
  robustness},
$\forall x, x'.\ \lVert x - x' \rVert<\epsilon \implies F(x')= F(x)$, which
checks that any two points that are similar (within a small acceptable
$\epsilon$) have the same label, this is not only inefficient to check but also fails to
hold for points on legitimate boundaries between regions.

Recent work uses Reluplex to check a more refined version of local and global robustness, where the confidence score of the lables of close inputs is checked to be within an acceptable parameter~\cite{KaBaDiJuKo17FVAV}. While this can potentially handle situations where the inputs are on the boundaries, it still requires manually finding the regions in which this check is likely to hold.

\paragraph{Our approach.}
We propose a novel technique, DeepSafe, for formally evaluating the robustness of deep
neural networks. The key notion underlying our approach is the use of a data-guided
methodology to determine regions that are likely to be safe (instead of
focusing on individual points). This enables characterizing the behavior of the network over partitions of the input space, which in turn makes the
network's behavior amenable to analysis and verification.

Our technique can automatically perform the following steps:
\begin{inparaenum}[(i)]
\item we propose a novel clustering algorithm to
    automatically partition the input domain into {\em regions} in
    which points are {\em likely} to have the same true label;
  \item these regions are then checked for robustness using an existing verification tool;
  \item the verification checks \emph{targeted robustness} which, given a
    specific incorrect label, guarantees that no input in the region is mapped by the NN to that label;
  \item for each region, the result of the targeted verification is either that
    the NN is {\em safe} (i.e. no points within the region are mapped to the
    target label), or, if it is not, an \emph{adversarial example}
    demonstrating that it is unsafe;
  \item robustness against all target labels (other than the correct label)
    indicates that the region is completely safe and all points within the
    region are mapped to the same correct label.
\end{inparaenum}

Thus, we {\em decompose} the robustness requirement for a NN into a number of
local proof obligations, one set for each region. Our approach provides several
benefits: we \emph{discover input regions that are likely to be robust} (this
is akin to finding likely invariants in program analysis) which are then
candidates for the safety checks; if a region is found to be safe, we provide
\emph{guarantees w.r.t all points within that region}, not just for individual
points as in previous techniques; and the discovered regions can \emph{improve
  the scalability of formal NN verification} by {\em focusing} the search for
adversarial examples only on the input space defined by a region. Note also
that the regions can be used to improve the scalability of the above-mentioned
approximate techniques (e.g gradient descent methods) by similarly restricting
the search to the input points confined by the region.

As the usual notion of safety might be too strong for many NNs, we introduce the concept of  \emph{targeted robustness}, analogous to targeted adversarial perturbations \cite{SzegedyZSBEGF13,GoodfellowSS14,FeinmanCSG17}. A region of the input space that is safe w.r.t. a specific target label indicates that within that region, the network is guaranteed to be robust against misclassification to the specific target label. Therefore, even if in that region the network is not completely robust against adversarial perturbations, we give guarantees that it is safe against specific targeted attacks.
As a simple example consider a NN used for perception in an autonomous car that classifies the images of a semaphore as red, green or yellow. We may want to guarantee that the NN will never classify  the image of a green light as a red light and vice versa but it may be tolerable to misclassify a green light as yellow, while still avoiding traffic violations.

The \textbf{\emph{contributions}} of our approach are as follows:

\begin{enumerate}
 \item \textbf{Label-guided clustering:} We present a clustering
   technique which is guided by the labels of the training data. This
   technique is an extension of the standard, unsupervised clustering algorithm
   \emph{kMeans}~\cite{KanungoMNPSW02} which we modified for our purposes.
   The output of the technique is a set
   of dense clusters within the input space, each of which contains
   training inputs that are close to each other (with respect to a
   given distance metric) and are known to share the same output
   label.

  \item \textbf{Well-defined safe regions:} Each cluster defines
    a subset of the inputs,
    a finite number of which belong
    to the training data. The key point is that the network
    is expected to display consistent behavior (i.e., assign the same
    label) over the entire cluster. Hence, the clusters can be
    considered as safe regions in which adversarial perturbations should
    not exist, backed by a data-guided rationale.
    Therefore, searching for adversarial perturbations within the
    cluster-based safe regions has a higher chance of producing
    \emph{valid} examples, i.e. inputs whose misclassification is
    considered erroneous network behavior.
    The size, location and boundaries of the clusters, which depend on the distribution of
    the training data can help guide the search:
    for instance, it often does not make sense to search for
    adversarial perturbations of inputs
    that lie on the boundaries between clusters belonging to different
    labels, as adversarial perturbations in those areas are likely to
    constitute acceptable network behavior.

  \item \textbf{Scalable verification:} Within each cluster, we use
    formal verification to prove that the network is robust or find
    adversarial perturbations. The verification of even simple neural
    networks is an NP-complete problem~\cite{KaBaDiJuKo17Reluplex},
    and is very difficult in practice. Focusing on clusters means that
    verification is applied to small input domains, making it more feasible
    and rendering the approach as a whole more scalable.
    Further, the verification of separate clusters can be done in
    parallel, increasing scalability even further.

  \item \textbf{Targeted robustness:} Our approach focuses on determining \emph{targeted safe regions}, analogous to targeted adversarial perturbations.
  A region of the input space that is safe w.r.t. a specific target label
  indicates that within that region, the network is guaranteed to not map any
  input to the specific target label. Therefore, even if the network is not
  completely robust against adversarial perturbations, we are able to give
  guarantees that it is safe against specific targeted attacks.

\end{enumerate}

Our proposed basic approach can have additional, interesting
applications. For example, clusters can be used in additional forms of
analysis, e.g. in determining whether the network is particularly susceptible to
specific kinds of perturbations in certain regions of the input
space. Also, the clustering approach is black-box in the sense
that it relies solely on the data distribution to determine the safe
regions (and not on the network parameters). Consequently, it is
applicable to a wide range of networks with various topologies and
architectures. Finally, as we later demonstrate, it is straightforward
to incorporate into the approach user-specific information (domain-specific constraints) regarding
which adversarial perturbations can be encountered in practice. This
helps guide the search towards finding valid perturbations.

The remainder of this paper is organized as follows. In
Section~\ref{sec:background}, we provide the needed background on
clustering, neural networks, and neural network verification. In
Section~\ref{sec:algorithm}, we describe in detail the steps of our
clustering-based approach, followed by an evaluation in
Section~\ref{sec:eval}. The possible limitations of our approach are discussed
in Section~\ref{sec:discussion}. Related work is then discussed in
Section~\ref{sec:related_Wrk}, and we conclude in Section~\ref{sec:conclusion}.


   \section{Background}
\label{sec:background}

\subsection{Clustering}
\label{sec:cls}

Clustering is an approach used to divide a population of data-points
into groups called clusters, such that the data-points in each cluster
are more similar (with respect to some metric) to other points in the
same cluster than to the rest of data-points.

Here we focus on a particularly popular clustering algorithm called
\emph{kMeans}~\cite{KanungoMNPSW02}  (although our approach could be implemented
using different clustering algorithms as well). Given a set of $n$
data-points $\{ x_{1} , \ldots , x_{n} \}$ and $k$ as the desired
number of clusters,  the algorithm partitions the points into $k$
clusters, such that the variance (also referred to as ``within cluster sum of squares'')
 within each cluster is minimal. The metric used to calculate the
 distance between points is customizable, and is typically the
 Euclidean distance ($L_2$ norm) or the Manhattan distance ($L_1$
 norm). For points $x_1=\langle x^1_1,\ldots,x^1_n\rangle$ and
 $x_2=\langle x^2_1,\ldots,x^2_n\rangle$ these are defined as:
\begin{equation}
\label{eq:norms}
\lVert x_1 - x_2 \rVert_{L_1} = \sum_{i=1}^n|x_i^1 - x_i^2|, \qquad
\lVert x_1 - x_2 \rVert_{L_2} = \sqrt{\sum_{i=1}^n(x_i^1 - x_i^2)^2 }
\end{equation}

The kMeans clustering is an iterative refinement algorithm which starts with $k$ random
points considered as the means (the \emph{centroids}) of $k$
clusters. Each iteration then comprises mainly of two steps:
\begin{inparaenum}[(i)]
\item assign each data-point to the cluster whose centroid is closest
  to it with respect to the chosen distance metric; and
\item re-calculate the new means of the clusters, which will serve as
  the new centroids.
\end{inparaenum}
The iterations continue until the assignment of data-points to
clusters does not change. This indicates that the clusters satisfy the
constraint that the variance within the cluster is minimal, and that the
data-points within each cluster are closer to each other than to
points outside the cluster.

\subsection{Neural Networks}
\label{sec:nn}
Neural networks and deep belief networks have been used in pattern
analysis, image classification, speech/audio recognition, perception
modules in self-driving cars so on. Typically, the objects in
such domains are high dimensional and the number of classes that the
objects need to be classified into is also high --- and so the
classification functions tend to be highly non-linear
over the input space. Deep learning operates with the
underlying rationale that groups of input parameters could be merged
to derive higher level abstract features, which enable the discovery of
a more linear and continuous classification function.
Neural networks are often used as \emph{classifiers}, meaning they
assign to each input an output label/class.
A neural network $F$ can thus be
regarded as a function that assigns to input $x$ an output label $y$,
denoted as $F(x) = y$.

Internally, a neural networks is comprised of multiple
layers of nodes called neurons, where each node refines and
extracts information from values computed by nodes in the previous layer.
A typical 3 layer neural network would consist of the following; the first layer is the \emph{input} layer, which takes
in the input variables (also called features) $x_1, x_2, \ldots,
x_n$. The second layer is a \emph{hidden} layer: each of its neurons
computes a weighted sum of the input variables using a unique weight
vector and a bias value, and then applies a non-linear
\emph{activation function} to the result. The last layer is the
\emph{output} layer, which uses the softmax function to make a
decision on the class for the input, based on the values computed by
the previous layer.

\subsection{Neural Network Verification}
Neural networks are trained and tested on finite sets of inputs and
outputs and are then expected to generalize well to previously-unseen
inputs. While this seems to work in many cases, when the network in
question is designed to be part of a safety-critical system, we may
wish to verify formally that certain properties hold for any possible
input. Traditional verification techniques often cannot directly be
applied to neural networks, and this has sparked a line of
work focused on transforming the problem into a format more amenable
to existing tools such as LP and SMT
solvers~\cite{Ehlers2017,HuangKWW17,PuTa10,PuTa12}.

While our approach is general in the sense that it could be coupled
with any verification technique, for evaluation purposes we used the
recently-proposed Reluplex
approach~\cite{KaBaDiJuKo17Reluplex}. Reluplex is a sound and complete
simplex-based verification procedure, specifically tailored to achieve
scalability on deep neural networks. Intuitively, the algorithm
operates by eagerly solving the linear constraints posed by the neural
network's weighted sums, while attempting to satisfy the non-linear
constraints posed by its activation functions in a
lazy manner. This often allows Reluplex to safely disregard many of
these non-linear constraints, which is where the bulk of the problem's
complexity stems from. Reluplex has been used in evaluating techniques
for finding and defending against adversarial perturbations~\cite{CaKaBaDi17}, and it has
also been successfully applied to a
real-world family of deep neural networks, designed to operate as
controllers in the next-generation Airborne Collision Avoidance System
for unmanned aircraft (ACAS Xu)~\cite{KaBaDiJuKo17Reluplex}.


   \Section{The DeepSafe Approach}
\label{sec:algorithm}

In this section we describe in greater detail the steps of our proposed approach:
\begin{inparaenum}[(i)]
\item clustering of training inputs;
\item cluster analysis;
\item cluster verification; and
\item processing of possible adversarial examples.
\end{inparaenum}

\subsection{Clustering of Training Inputs}

\begin{figure}[t]
\begin{tabular}{c c}
\begin{minipage}[t]{.45\columnwidth}
{\footnotesize
\begin{CodeOut}
\begin{verbatim}
function rep = next_fun_acas(p)
movefile('count.csv','countin.csv');
n1in = importdata("countin.csv");
m = size(n1in);
rep = 0;
for sn = 1 : m(1)
    num = n1in(sn,1);
    lo = "cluster"+ num2str(num)
    +".csv";
    ln = "clusterin"+num2str(num)
    +".csv";
    loc = char(lo);
    lnc = char(ln);
    movefile(loc,lnc);
end
cnt = 0;
for yn = 1 : m(1)
    X = importdata("clusterin"+
          num2str(n1in(yn,1))+".csv");
    [idx,C] = kmeans(X,n1in(yn,2),
          'Distance','sqeuclidean');
    Xn = [X,idx];
    Yn = sortrows(Xn,7);
    Un = unique(Yn(:,7));
    nn = size(Un);
\end{verbatim}
\end{CodeOut}
}
\end{minipage}
&
\begin{minipage}[t]{.45\columnwidth}
{\footnotesize
\begin{CodeOut}
\begin{verbatim}
for xn = 1 : nn(1)
  Zn = Yn(Yn(:,7) == Un(xn,1),1:6);
  Bn = unique(Zn(:,6));
  n0n = size(Bn);
  wn = xn + cnt;
  if (n0n(1,1) > 1)
    rep = 1;
    csvwrite("cluster"+num2str(wn)
        + ".csv",Zn);
    n1n(1,1) = wn;
    n1n(1,2) = n0n(1,1);
    dlmwrite("count.csv",n1n,
          '-append','delimiter',',');
  else
    sn = Un(xn,1);
    csvwrite("clusterFinal"+num2str(wn)
        + "_" + num2str(p)+".csv",C(sn,:));
    dlmwrite("clusterFinal"+num2str(wn)
        + "_" + num2str(p)+".csv",Zn,
          '-append','delimiter',',');
  end
end
cnt = wn;
end end
\end{verbatim}
\end{CodeOut}
}
\end{minipage}\\
\multicolumn{2}{c}{}
\end{tabular}
\Caption{\footnotesize{MATLAB code for modified kMeans clustering. \label{fig:psd}}}
\end{figure} 

\begin{figure}[H]
\begin{center}
\begin{tabular}{c c}
\begin{minipage}[t]{.45\columnwidth}
\includegraphics[scale=0.3]{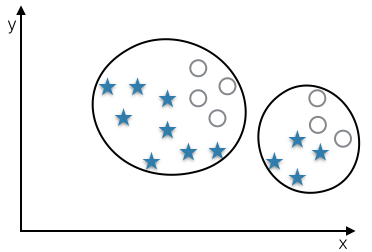}
\end{minipage}
&
\begin{minipage}[t]{.45\columnwidth}
\includegraphics[scale=0.3]{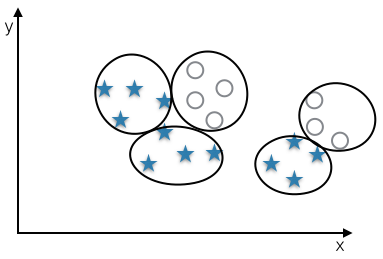}
\end{minipage}\\
\multicolumn{2}{c}{}
\end{tabular}
\Caption{\footnotesize{a) Original clusters with k=2, b) Clusters with modified kMeans \label{fig:clusters}}}
\end{center}
\end{figure} 
\Comment{
\begin{minipage}{\linewidth}
      \begin{minipage}{0.45\linewidth}
          \begin{figure}[H] \label{fig:original}
\includegraphics[scale=0.3]{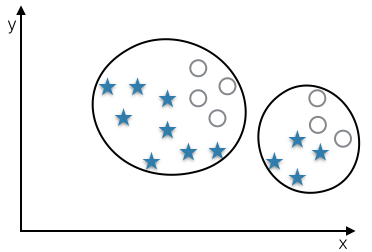}
\caption{Original clusters with k=2}
          \end{figure}
      \end{minipage}
      \hspace{0.05\linewidth}
      \begin{minipage}{0.45\linewidth}
          \begin{figure}[H] \label{fig:mod}
\includegraphics[scale=0.3]{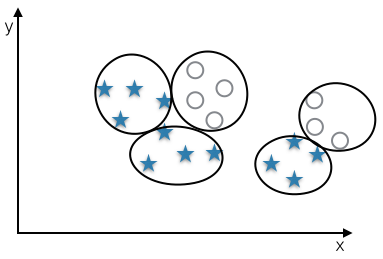}
\caption{Clusters with modified kMeans}
          \end{figure}
      \end{minipage}
\vspace{0.5cm}
  \end{minipage}
}

In our approach, we use the kMeans clustering algorithm (see Section~\ref{sec:cls})
to perform clustering over the training inputs. By training inputs, we mean
{\em all} inputs whose correct labels or output classes are known:
this includes the training, validation, and test sets.
We note that our technique works even in the absence of the training data, e.g. by applying the clustering
to a set of randomly generated inputs that are labeled according to a given trained network. The user will
then need to check that the labels are valid.

The kMeans approach is typically
an unsupervised technique, meaning that clustering is based purely on the
similarity of the data-points themselves, and does not depend on their
labels. Here, however, we use the labels to guide the clustering
algorithm into generating clusters that have consistent labeling (in
addition to containing points that are similar to each other).
The modified clustering algorithm starts by setting the number of
clusters $k$, which is an input to the kMeans algorithm, to be equal
to the number of unique labels. Once the clusters are obtained, we
check whether each cluster contains only inputs with the same label.
kMeans is then applied again on each cluster that is found to contain
multiple labels, with $k$ set to the number of unique labels within
that cluster. This effectively breaks the ``problematic'' cluster into
multiple sub-clusters.
The process is repeated until all clusters contain inputs which share
a single label. The number of clusters, which is an input parameter of
the kMeans algorithm, is often chosen arbitrarily. In our approach, we
take guidance from the training data to customize the number of
clusters to the domain under consideration. Pseudocode for the MATLAB
implementation of the algorithm appears in Fig.~\ref{fig:psd}.
In every iteration, {\tt count.csv} maps the index of each input dataset, {\tt clusterINDEX.csv}, to the number of unique labels its instances correspond to.  Clusters whose instances correspond to the same label are named {\tt clusterFinalINDEX.csv}.

Let us consider a toy example with training data labeled as either stars or circles. Each training data point is characterised by two dimensions/attributes (x,y). The original kMeans algorithm with $k = 2$, will partition the training inputs into 2 groups, purely based on proximity w.r.t. the 2 attributes (Fig.~\ref{fig:clusters}a). However, this groups stars and circles together. Our modified algorithm creates the same partitions in its first iteration, however, since each cluster does not satisfy the invariant that it only contains training inputs with the same label, it proceeds to iteratively divide each cluster into two sub-clusters until this invariant is satisfied. This creates 5 clusters as shown in (Fig.~\ref{fig:clusters}b); 3 with label star and 2 with label circle. This example is typical
for domains such as image classification, where even a
small change in some attribute values for certain inputs could change
their label.

\sloppy
Our modified clustering algorithm typically produces small, dense
clusters of consistently-labeled inputs. The underlying assumption of
our approach is that each such cluster therefore constitutes a
\emph{safe region} in which all inputs (and not just the training
points) should be labeled
consistently. For instance, in our toy example, the algorithm creates relatively small clusters of stars and circles, separating them from each other. Searching within these clusters, which represent small neighborhoods of inputs belonging to the same label, may yield more meaningful results than searching regions within an arbitrary distance of each input.

Each of the clusters generated
by kMeans is characterized by a centroid $\cen{}$ and a radius $R$, indicating
the maximum distance of any instance in the cluster from its
centroid. While inputs deep within each cluster are expected to be
labeled consistently, the boundaries of the clusters may lie in
low-density regions, and could have different labels. In order to
improve the accuracy of our approach, we shrink the clusters by
replacing $R$ with the average distance
of any instance from the centroid, denoted $r$.
This increases the likelihood that all points within the cluster
should be consistently labeled, and that any deviation from this would
constitute a valid adversarial perturbation. The training inputs within a cluster have the same label,
which we refer to as the label of the cluster, $l$. To summarize, the
main hypothesis behind our approach is:
\begin{hypothesis}
\label{mainHypothesis}
 For a given cluster $C$, with centroid $\cen{}$ and radius $r$, any
 input $x$ within distance $r$ from $\cen{}$ has the same true label
 $l$ as that of the cluster:

\[
\lVert x - \cen{} \rVert \leq r \quad\Rightarrow\quad label(x) = l
\]
\end{hypothesis}
It immediately follows from this hypothesis that any point $x'$ in the
cluster which is assigned a different label $F(x')\neq l$ by the
network constitutes an adversarial perturbation. An adversarial input for our toy example could be as follows; a hypothetical NN, in its process of input transformation may incorrectly bring some of the inputs within the cluster with stars close to circles, thereby classifying them as circle.

\paragraph{Distance metric.}
The similarity of the inputs within a cluster is determined by the
distance metric used for calculating the proximity of the
inputs. Therefore, it is important to choose a distance metric that
generates acceptable levels of similarity for the domain under
consideration. Our approach assumes that every input, characterized by
$m$ attributes, can be considered as a point in Euclidean space. The
Euclidean distance (Eq.~\ref{eq:norms}) is a commonly used metric
for measuring the proximity of points in Euclidean space. However,
recent studies indicate that the usefulness of Euclidean distance
in determining the proximity between points diminishes as the
dimensionality increases ~\cite{AggarwalHK01}. The Manhattan distance
(Eq.~\ref{eq:norms}) has been found to capture proximity more
accurately at high dimensions. Therefore, in our experiments, we set
the distance metric depending on the dimensionality of
the input space (Section~\ref{sec:eval}). Other distance metrics can be easily accommodated by our approach.

\subsection{Cluster Analysis}

The clusters obtained as a result of the previous step characterize
the behavior of the network over large chunks of the input
space. Analysis of these clusters can provide useful insights
regarding the behavior and accuracy of the network. Listed below are cluster properties that we use
in our approach. 

\begin{itemize}
  \item \textbf{Density:} We define the density of a cluster to be the
    number of instances per unit distance within the cluster. For a
    cluster with $n$ points and average distance $r$ from the points
    to the centroid, the cluster's density is defined to be $n/r$. A
    cluster with high density contains a large number of instances at
    close proximity to each other (and because of our modified
    clustering algorithm, these instances are also labeled
    consistently). We assume that in such clusters
    Hypothesis~\ref{mainHypothesis} holds, since it seems undesirable
    that the network should assign a different label to inputs that
    lie within the average distance from the centroid. However, this
    cannot be said regarding a cluster with low density, which
    encompasses fewer inputs belonging to the same label and which are spread out.
    Therefore, for the purpose of the robustness check, we disregard
    clusters with low density, in order to increase the chances of
    examining only valid safe regions, and hence of detecting only
    valid adversarial perturbations.

  \item \textbf{Centroid behavior:} The centroid of a cluster
    can be considered as a representative for the
    behavior of the network over that cluster --- especially when the
    cluster's density is high. A classifier neural network assigns to
    each input a score for each possible
    label, representing the network's level of confidence that the
    input should have that label (the label with the highest confidence
    is the one then assigned to the input).\Comment{ Observing the levels of
    confidence assigned to each label for the centroid can provide
    hints as to the existence of \emph{targeted adversarial
      perturbations}: perturbations that seek not just to misclassify
    an input, but to cause it to be classified as a specific
    label~\cite{}.}
    Intuitively, targeted adversarial perturbations are
    more likely to exist, e.g., for the label whose level of
    confidence at the centroid is
    the second highest, than for the least likely label. In our
    approach, we use the label scores to look for \emph{targeted
      safe
      regions}: regions of the input space within which the network
  is guaranteed to be robust against misclassification to a specific
  target label.
\end{itemize}

\subsection{Cluster Verification}
Having identified and analyzed the clusters, we next use the Reluplex
tool~\cite{KaBaDiJuKo17Reluplex} to verify a formula representing
the negation of Hypothesis~\ref{mainHypothesis}. This is done in a
\emph{targeted manner} or on a per label basis. If the negated
hypothesis is shown not to hold,
the region is indeed safe with respect to that label
(\emph{targeted safe region}); and otherwise, Reluplex
provides a
satisfying assignment, which constitutes a valid adversarial
perturbation. The encoding is shown in Eq.~\ref{eq:relform}:
\Comment{
\begin{equation}\label{eq:relform}
    \bigvee_{l'\in L-\{l\}}
    \left(
  \exists x.\quad
    \lVert x - \cen{} \rVert_{L_1} \leq r \quad\bigwedge\quad
 \score(x,l') \geq \score(x,l)
    \right)
\end{equation}}

\begin{equation}\label{eq:relform}
  \exists x.\quad
    \lVert x - \cen{} \rVert_{L_1} \leq r \quad\wedge\quad
 \score(x,l') \geq \score(x,l)
\end{equation}
Here, $x$ represents an input point, and $\cen{}$, $r$
and $l$ represent the centroid, radius and label of the cluster,
respectively. $l'$ represents a label other than $l$.
Reluplex models the network without the final softmax
layer, and so the networks' outputs correspond to the levels of
confidence that the network assigns to each of the possible labels; we
use $\score(x,y)$ to denote the level of confidence assigned to label
$y$ at point $x$. Intuitively,  the formula holds for a given $l'$ if and only if there
exists a point $x$
within distance at most $r$ from $\cen{}$, for which $l'$
 is assigned higher confidence than $l$. Consequently, if the property
 does \emph{not} hold, then for every $x$ within the cluster $l$ has
 its score higher than $l'$. This ensures targeted robustness of the
 network for label $l'$: the network is guaranteed not to misclassify
 any input within the region to the target label $l'$.  Note that for overlapping clusters with different labels, there is uncertainty regarding the desired labels for the clusters' intersection. Our method of reducing the clusters' radius can serve to exclude such regions.

The property in Eq.~\ref{eq:relform} is checked sequentially for
every possible $l'\in L-\{l\}$, where $L$ denotes the set of all
possible labels. If the property is unsatisfiable for all $l'$,
it ensures \emph{complete robustness of the
  network for inputs within the cluster}; i.e., the network is guaranteed
not to misclassify any input within the region to any other label.
This can be expressed more formally as shown below:
\begin{equation}\label{eq:safe-region}
   \forall  x.\ \ \lVert x - \cen{} \rVert_{L_1} \leq r
   \quad\Rightarrow\quad \forall l\in L-\{l\}.\quad \score(x,l)\geq \score(x,l')
\end{equation}
from which it follows that Hypothesis~\ref{mainHypothesis} holds,
i.e. that:
\begin{equation} \label{eq:com-safe-region}
  \forall x.\ \ \lVert x - \cen{} \rVert_{L_1} \leq r \quad\Rightarrow\quad label(x) = l
\end{equation}

As is the case with many SMT-based solves, Reluplex typically solves
satisfiable queries more quickly that unsatisfiable ones.
 Therefore, in order to optimize performance we test the possible
 target labels $l'$ in descending order of the scores that they are
 assigned at the centroid,  $\score(\cen{},l')$. Intuitively, this is
because the label with the 2$^{nd}$-highest score is more likely to yield
a satisfiable query, etc.

\paragraph{Distances in Reluplex.}
Reluplex takes as input a conjunction of linear equations and certain
piecewise-linear constraints. Consequently, it is straightforward to model the neural
network itself and the query in Eq.~\ref{eq:relform}. Our ability to
encode the distance constraint from that equation, $\lVert x-
\cen{}\rVert\leq r$, depends on the distance metric being used. While
$L_1$ is piecewise linear and can be encoded, $L_2$ unfortunately cannot.

When dealing with domains where $L_2$ distance is a better measure of
proximity, we thus use the following approximation. We perform the
clustering phase using the $L_2$ distance metric as described before,
and for each cluster obtain the radius $r$. When verifying the
property in Eq.~\ref{eq:relform}, however, we use the $L_1$ norm. Because
$\lVert x- \cen{}\rVert_{L_1}\leq \lVert x- \cen{}\rVert_{L_2}$, it is
guaranteed that the verification is conducted within the cluster under
consideration, and any adversarial perturbation discovered will thus
be valid. If the verification shows that the network is robust,
however, then this holds only for the portion of the cluster that was
checked. This limitation can be waived by using a verification
technique that directly supports $L_2$, or by enhancing Reluplex to
support it.
\Comment{
\begin{figure}[t]
\begin{center}
\includegraphics[width=4.0in]{figures/clusters.pdf}
\vspace{-2cm}
\Caption {\footnotesize{Analysis coverage within clusters} \label{fig:cov_cls}}
\end{center}
\end{figure}
}
\paragraph{Clusters and scalability.}
The main source of computational complexity in neural network
verification is the presence of non-linear, non-convex activation
functions. However, when restricted to a small sub-domain of the input
space, these functions
may present purely linear behavior --- in which case they can be
disregarded and replaced with a linear constraint, which greatly
simplifies the problem.
Consequently, performing verification within small domains is
beneficial, as many activation functions
can often be disregarded. Our approach naturally involves
verification queries on small clusters, which tends to be very helpful
in this regard. Reluplex has built-in \emph{bound
  tightening}~\cite{KaBaDiJuKo17Reluplex} functionality to detect
such cases; and we leverage this functionality by computing lower and
upper bounds for each of the input variables within the cluster, and
provide these as part of our query to Reluplex.
\Comment{ Figure Fig.~\ref{fig:cov_cls} illustrates the reduced search space
for our example clusters, with the regions in black indicating those
parts of each cluster that are trimmed by restricting variable
ranges to those that actually appear in the cluster.}

Our approach lends itself to more scalable verification also through
parallelization. Because each cluster involves stand-alone
verification queries, their verification can be performed in
parallel. Also, because Eq.~\ref{eq:relform} is checked independently for
every $l'$, these queries can also be
performed in parallel --- expediting the process even further~\cite{KaBaDiJuKo17FVAV}.

\subsection{Processing Possible Adversarial Perturbations}
A SAT  solution to Eq.~\ref{eq:relform} for any target label ($l'$)
indicates the presence of an input within the region for which the
network assigns a higher score to label $l'$ than to $l$. The check
for the validity of the adversarial example needs to be done by the user/domain
expert. Note that this does not mean that $l'$ has the highest score;
i.e. it need not be a targeted adversarial example for $l'$. In some cases,
there may be specific constraints on inputs that can be considered
valid adversarial examples. We have been able to successfully model such
domain-specific constraints for ACAS Xu to generate valid adversarial perturbations (Section~\ref{sec:eval}).


   \section{Case Studies}
\label{sec:eval}
We implemented DeepSafe using MATLAB R2017a for the clustering algorithm and Reluplex v1.0 for verification. The runs were dispatched on a 8-Core 64GB server running Ubuntu 16.0.4. We evaluated DeepSpace on  two case studies. The first network is part of a
real-world controller for the next-generation Airborne Collision Avoidance System for unmanned aircraft (ACAS Xu), a highly safety-critical system. The second network is a digit classifier over the popular MNIST image dataset.
\Comment{
\begin{enumerate}
  \item \textbf{RQ1:} How useful is DeepSafe in providing guarantees about the behavior of the network over the input space?
  \item \textbf{RQ2:} Can the computed regions and their respective labels be used as sound oracles for evaluating the network?
  \item \textbf{RQ3:} Is there a correlation between properties of clusters (such as density and the confidence of the centroid's label) and the likelihood of the corresponding regions being safe?
\end{enumerate}
}

\begin{table}
\centering
\caption{ Summary of the analysis for the ACAS Xu network for 210 clusters}
\begin{tabular}{|c|c|c|c|c|c|}
\hline
\multicolumn{1}{|c|}{property} & \multicolumn{1}{|c|}{\# clusters} & \multicolumn{1}{|c|}{min radius} & \multicolumn{1}{|c|}{time(hours)} & \multicolumn{1}{|c|}{\#queries} \\
\hline
\multicolumn{1}{|c|}{safe} & \multicolumn{1}{|c|}{125} & \multicolumn{1}{|c|}{0.084} & \multicolumn{1}{|c|}{4} & \multicolumn{1}{|c|}{11.8} \\
\hline
\multicolumn{1}{|c|}{targeted safe} & \multicolumn{1}{|c|}{52} & \multicolumn{1}{|c|}{0.135} & \multicolumn{1}{|c|}{7.6} & \multicolumn{1}{|c|}{14.4} \\
\hline
\multicolumn{1}{|c|}{time out} & \multicolumn{1}{|c|}{33}& \multicolumn{1}{|c|}{NA} & \multicolumn{1}{|c|}{12} & \multicolumn{1}{|c|}{NA} \\
\hline
\end{tabular}
\label{table:acasXuSum}
\end{table}

\begin{table}
\centering
\caption{
Details of the analysis for some clusters for ACAS Xu
}
\begin{tabular}{|c|c|c|c|c|c|}
\hline
\multicolumn{1}{|c|}{cluster\#} & \multicolumn{1}{|c|}{safe} & \multicolumn{1}{|c|}{radius}  & \multicolumn{1}{|c|}{\#queries} & \multicolumn{1}{|c|}{time(min)}& \multicolumn{1}{|c|}{slice} \\
\multicolumn{1}{|c|}{} & \multicolumn{1}{|c|}{for label} & \multicolumn{1}{|c|}{}  & \multicolumn{1}{|c|}{} & \multicolumn{1}{|c|}{}& \multicolumn{1}{|c|}{(Y/N)} \\
\hline
\multicolumn{1}{|c|}{5282} & \multicolumn{1}{|c|}{1} & \multicolumn{1}{|c|}{0.04}  & \multicolumn{1}{|c|}{1} & \multicolumn{1}{|c|}{5.45}& \multicolumn{1}{|c|}{N} \\ \cline{2-6}
\multicolumn{1}{|c|}{label:0} & \multicolumn{1}{|c|}{2} & \multicolumn{1}{|c|}{0.04}  & \multicolumn{1}{|c|}{1} & \multicolumn{1}{|c|}{3.91}& \multicolumn{1}{|c|}{N} \\ \cline{2-6}
\multicolumn{1}{|c|}{} & \multicolumn{1}{|c|}{3} & \multicolumn{1}{|c|}{0.04}  & \multicolumn{1}{|c|}{1} & \multicolumn{1}{|c|}{3.57}& \multicolumn{1}{|c|}{N} \\ \cline{2-6}
\multicolumn{1}{|c|}{} & \multicolumn{1}{|c|}{4} & \multicolumn{1}{|c|}{0.04}  & \multicolumn{1}{|c|}{1} & \multicolumn{1}{|c|}{4.01}& \multicolumn{1}{|c|}{N} \\
\hline
\multicolumn{1}{|c|}{1783} & \multicolumn{1}{|c|}{1} & \multicolumn{1}{|c|}{0.16}  & \multicolumn{1}{|c|}{4} & \multicolumn{1}{|c|}{1.28}& \multicolumn{1}{|c|}{Y} \\ \cline{2-6}
\multicolumn{1}{|c|}{label:0} & \multicolumn{1}{|c|}{2} & \multicolumn{1}{|c|}{0.17}  & \multicolumn{1}{|c|}{1} & \multicolumn{1}{|c|}{279}& \multicolumn{1}{|c|}{N} \\ \cline{2-6}
\multicolumn{1}{|c|}{} & \multicolumn{1}{|c|}{3} & \multicolumn{1}{|c|}{0.17}  & \multicolumn{1}{|c|}{1} & \multicolumn{1}{|c|}{236}& \multicolumn{1}{|c|}{N} \\ \cline{2-6}
\multicolumn{1}{|c|}{} & \multicolumn{1}{|c|}{4} & \multicolumn{1}{|c|}{0.17}  & \multicolumn{1}{|c|}{1} & \multicolumn{1}{|c|}{223}& \multicolumn{1}{|c|}{N} \\
\hline
\multicolumn{1}{|c|}{2072} & \multicolumn{1}{|c|}{0} & \multicolumn{1}{|c|}{0.06}  & \multicolumn{1}{|c|}{1} & \multicolumn{1}{|c|}{11.51}& \multicolumn{1}{|c|}{N} \\ \cline{2-6}
\multicolumn{1}{|c|}{label:1} & \multicolumn{1}{|c|}{2} & \multicolumn{1}{|c|}{0.014}  & \multicolumn{1}{|c|}{9} & \multicolumn{1}{|c|}{0.98}& \multicolumn{1}{|c|}{N} \\ \cline{2-6}
\multicolumn{1}{|c|}{} & \multicolumn{1}{|c|}{3} & \multicolumn{1}{|c|}{0.011} & \multicolumn{1}{|c|}{7} & \multicolumn{1}{|c|}{0.71}& \multicolumn{1}{|c|}{N} \\ \cline{2-6}
\multicolumn{1}{|c|}{} & \multicolumn{1}{|c|}{4} & \multicolumn{1}{|c|}{0.012} & \multicolumn{1}{|c|}{5} & \multicolumn{1}{|c|}{0.58}& \multicolumn{1}{|c|}{N} \\
\hline
\multicolumn{1}{|c|}{6138} & \multicolumn{1}{|c|}{1} & \multicolumn{1}{|c|}{0.089} & \multicolumn{1}{|c|}{9} & \multicolumn{1}{|c|}{103.2}& \multicolumn{1}{|c|}{N} \\ \cline{2-6}
\multicolumn{1}{|c|}{label:0} & \multicolumn{1}{|c|}{2} & \multicolumn{1}{|c|}{0.11}  & \multicolumn{1}{|c|}{4} & \multicolumn{1}{|c|}{2.86}& \multicolumn{1}{|c|}{N} \\
\hline
\end{tabular}
\label{table:acasXuDet}
\end{table}

\subsection{ACAS Xu}
ACAS X is a family of collision avoidance systems for aircraft which
is currently under development by the Federal Aviation Administration (FAA)~\cite{ACASXU}.
ACAS Xu is the version for unmanned aircraft control. It is intended
to be airborne and receive sensor information regarding the drone (the
\emph{ownship}) and
any nearby intruder drones, and then issue horizontal turning
advisories aimed at preventing collisions. The input sensor data includes:
\begin{inparaenum}[(i)]
\item $\rho$: distance from ownship to intruder;
\item $\theta$: angle of intruder relative to ownship heading direction;
\item $\psi$: heading angle of intruder relative to ownship heading direction;
\item $v_{\text{own}}$: speed of ownship;
\item $v_{\text{int}}$: speed of intruder;
\item $\tau$: time until loss of vertical separation; and
\item $a_{\text{prev}}$: previous advisory.
\end{inparaenum}
The five possible output
actions are as follows: Clear-of-Conflict (COC), Weak Right, Weak
Left, Strong Right, and Strong Left. Each advisory is assigned a score,
with the lowest score corresponding to the best action. The FAA is currently exploring an implementation of ACAS Xu that uses
an array of 45 deep neural networks. These networks were obtained by
discretizing the two parameters, $\tau$ and $a_{\text{prev}}$, and
so each network contains five input dimensions and treats
$\tau$ and $a_{\text{prev}}$ as constants. Each network
has 6 hidden layers and a total of 300 hidden ReLU activation
nodes.

We applied our approach to several of the ACAS XU networks. We describe here in detail the results for one network.
Each input consists of 5 dimensions and is assigned one of 5 possible
output labels, corresponding to the 5 possible turning advisories for the drone (0:COC, 1:Weak Right, 2:Weak
Left, 3:Strong Right, and 4:Strong Left). We were supplied a set of cut-points, representing valid important values for each dimension, by the domain experts~\cite{ACASXU}. We generated 2662704 inputs (cartesian product of the values for all the dimensions). The network was executed on these inputs and the output advisories (labels) were verified. These were considered as the inputs with known labels for our experiments.

The labeled-guided clustering algorithm was applied on the inputs using the $L_2$
distance metric.  Clustering yielded 6145 clusters with more than one input and 321
single-input clusters. The clustering took 7 hours. For each cluster we computed a region, characterized by a centroid (computed by kMeans), radius (average distance of every cluster instance from the centroid), and the expected label (the label of all the cluster instances).

We first evaluated the network on all the centroids as they are considered representative of the entire cluster and should ideally have the expected label. The network assigned the expected label for the centroids of 5116 clusters (83\% of total number of clusters). For the remaining 1029 clusters,
we found that they contained few labeled instances spread out in large areas.
Therefore, we considered these clusters were not precise and our analysis was inconclusive.
For singleton clusters, we fall back to checking local robustness using previous techniques~\cite{KaBaDiJuKo17Reluplex}.  These stand-alone points serve to identify portions of the input space which require more training data, thus potentially more vulnerable to adversarial perturbations.

Amongst the remaining 5116 clusters, we picked randomly 210 clusters to illustrate our technique. These clusters contain 659315 labeled inputs (24\% of the total inputs with known labels).  For each region corresponding to the respective clusters, we applied DeepSafe to check equation~\ref{eq:relform} for every label. The distance metric used was $L_1$ since $L_2$ can not be handled by Reluplex (see section~\ref{sec:algorithm} that explains why this is still safe).
The results are presented in tables ~\ref{table:acasXuSum} and~\ref{table:acasXuDet}.
The \emph{min radius} in table ~\ref{table:acasXuSum}, refers to the average minimum radius around the centroid of each region for which the safety guarantee applies (averaged over the total number of regions for that safety type).
The \emph{\# queries} refers to the number of times the solver had to be invoked until an UNSAT was obtained, averaged over all the regions for that property.


DeepSafe was able to identify 125 regions which are completely \emph{safe}, i.e. the network yields a label consistent with the neighboring labeled inputs within the region. 52 regions are \emph{targeted safe}, the network is safe against misclassifying inputs to certain labels. For instance, the inputs within region 6138 (Table~\ref{table:acasXuDet}) with an expected label 0 (COC), were safe against misclassification only to labels 1 (weak right) and 2 (weak left). The solver timed out without returning any result for the remaining labels. The analysis timed out without returning a concrete result for any label for 33 clusters. A time out does not allow to provide a proof for the regions, although the likely answer is safe (generally, solvers take much longer when there is no solution).

The \emph{min radius} in table ~\ref{table:acasXuSum}, refers to the average minimum radius around the centroid of each region for which the safety guarantee applies (averaged over the total number of regions for that safety type).

The \emph{\# queries} refers to the number of times the solver had to be invoked until an UNSAT was obtained, averaged over all the regions for that property.

\Comment{
\noindent{\textbf{Discretization to estimate input space coverage:}}
\textbf{Instead of this, we could perhaps state the number of labeled inputs enclosed within the safety regions and use that as a measure of input space coverage?}
The \emph{average \# inputs} refers to an estimation of the total number of discrete inputs that we give guarantee for. The number of input values covered by the solver is infinite, considering that each dimension has continuous numeric real values. However, to get an estimate of the \% of all possible inputs that DeepSafe has been able to provide guarentee for, we adopt the following approximate discretization technique. We looked at the range of values for each dimension and the minimum difference between distinct values for each dimension amongst the 2662704 inputs with known labels. For the first dimension $\rho$, the minimum difference between subsequent normalized values used for training was 0.01, for the second and third dimensions it was 0.05 and the 0.1 for the fourth and fifth dimensions respectively. We considered hypothetically that there could be 10 possible valid values between subsequent values (with known labels). Therefore we discretized the first dimension with a step-size of 0.001, the second and third with 0.005 and fourth and fifth with 0.01 respectively. This yields a total of $4$ X $10^{11}$ distinct inputs in the entire input space. We calculated the number of such discrete inputs that could be enclosed within a cluster, given its radius.}

\subsection{MNIST Image Dataset}
The MNIST database is a large collection of handwritten digits that is
commonly used for training various image processing systems~\cite{MNIST}. The dataset has 60,000 training input images, each characterized
by 784 attributes and belonging to one of 10 labels. We used a network that
comprised of 3 layers, each with 10 ReLU activation nodes. Clustering
was applied using the $L_1$ distance metric. It yielded 6654 clusters
with more than one input and 5681 single-input clusters.
The clustering consumed 10 hours.
A separate process for verification of each cluster was spawned with a time-out of 12 hours.

\begin{table}
\centering
\caption{Summary of the analysis for MNIST network for 80 clusters}
\begin{tabular}{|c|c|c|c|c|}
\hline
\multicolumn{1}{|c|}{property} & \multicolumn{1}{|c|}{\# clusters}  & \multicolumn{1}{|c|}{min radius} & \multicolumn{1}{|c|}{time(hours)} & \multicolumn{1}{|c|}{\# queries} \\
\hline
\multicolumn{1}{|c|}{safe} & \multicolumn{1}{|c|}{7} & \multicolumn{1}{|c|}{2.46} & \multicolumn{1}{|c|}{11.27} & \multicolumn{1}{|c|}{2.85} \\
\hline
\multicolumn{1}{|c|}{targeted safe} & \multicolumn{1}{|c|}{63}& \multicolumn{1}{|c|}{5.19} & \multicolumn{1}{|c|}{11.02} & \multicolumn{1}{|c|}{4.87} \\
\hline
\multicolumn{1}{|c|}{time out} & \multicolumn{1}{|c|}{10}& \multicolumn{1}{|c|}{NA} & \multicolumn{1}{|c|}{12} & \multicolumn{1}{|c|}{NA} \\
\hline
\end{tabular}
\label{table:MNISTSum}
\end{table}

For the singleton clusters, as is the case with ACAS Xu, we performed local robustness checking as in previous approaches.

Table~\ref{table:MNISTSum} shows the summary of the results for the runs for 80 clusters that we selected for evaluation. In past studies, the MNIST network has been shown to be extremely vulnerable to misclassification on adversarial perturbations even with state-of-the art networks~\cite{Carlini017}. Therefore, as expected, it is easy to determine SAT solutions and they were discovered very fast (within a minute). However, it is very time consuming to prove safety; the verification time is much higher than that of the ACAS Xu application as it is mainly impacted by the large number of input variables (784 attributes). We would like to highlight that our work is the first to successfully identify safety regions for MNIST even on a fairly vulnerable network.

For 7 clusters, the solver returned UNSAT for all labels within 12 hours. For 30 clusters, the solver returned  UNSAT only for few labels but timed out before returning any solution for the other labels. These have been included in the \emph{targeted safe} property in the table.  Additionally, based on the nature of this domain, we can consider it safe to assume that if for any label the solver does not return a SAT solution within 10 hours, then it is safe w.r.t. that label even if it does not prove unsatisfiability within this time. This happened to be the case for 33 clusters, where the solver could not find a solution for a specific target label despite executing for more than 10 hours. These have been included in the \emph{targeted safe} type as well. For 10 of the remaining clusters, the solver kept finding adversarial examples despite iterative reductions of the radius and the time-out occurred before the radius reduced to 0. These have been included as time out in the table, since we cannot determine for sure if the region should be marked unsafe for the specific labels.

    \section{Threats to Validity}
\label{sec:discussion}
\label{sec:dis}
We discuss below possible threats to the validity of our approach and the experiments.

\begin{itemize}
  \item \textbf{Invalid adversarial examples:} The validity of an adversarial perturbation depends on the accuracy of the cluster in identifying regions of the input space which should ideally be given the same label. Clustering as used in our approach typically generates small dense groups or small neighborhoods around known inputs. The NN, on the other hand, attempts to abstract the input by focusing on certain features more than the others in order to be able to assign a unique label to it. This abstraction also enables generalization of the network to other inputs not part of the training data. However, this process tends to make the network inaccurate on inputs in close neighborhoods to known inputs. Therefore, although clustering cannot be considered as an alternative classifier to NN for any input, it can be considered to be accurate or an oracle in close neighborhoods of known inputs. This can however be impacted by the presence of noise in the input space due to irrelevant attributes, which the NN sieves out.

  \item \textbf{Invalid safety regions:} There could be a scenario where both the cluster and the network agree on the labels for all inputs within a region, however, ideally some of them need to be classified to a different label. This can happen when the training data is not representative enough.

  \item \textbf{Generalization of experimental results:} The current implementation of the solver Reluplex used in our prototype tool only supports piecewise-linear activation functions. This could limit the generalization of the experimental results to networks which use other types of activation functions.

\end{itemize}

   \section{Related Work}
\label{sec:related_Wrk}

The vulnerability of neural networks to adversarial perturbations was
first discovered by Szegedy et. al. in 2013~\cite{SzegedyZSBEGF13}. They model the problem of
finding the adversarial example as a constrained minimization
problem. Goodfellow et al.~\cite{GoodfellowSS14} introduced the Fast Gradient Sign Method for crafting
adversarial perturbations using the derivative of the model’s loss
function with respect to the input feature vector. They show that NNs
trained for the
MNIST and CIFAR-10 classification tasks can be fooled with a high success
rate. An extension of this approach applies the technique in an
iterative manner~\cite{KurakinGB16a}.  Jacobian-based Saliency Map Attack (JSMA)~\cite{PapernotMJFCS16} proposed a method
for targeted misclassification by exploiting the forward derivative
of a NN to find an adversarial perturbation that will force the model
to misclassify into a specific target class. Carlini et. al.~\cite{Carlini017}
recently proposed an approach that could not be resisted by
state-of-the-art networks such as those using defensive
distillation. Their optimization algorithm uses better loss functions
and parameters (empirically determined) and uses three different
distance metrics.

The DeepFool~\cite{Moosavi-Dezfooli16} technique simplifies the domain by considering
the network to be completely linear. They compute adversarial inputs
on the tangent plane (orthogonal projection) of a point on the
classifier function. They then introduce non-linearity to the model,
and repeat this process until a true adversarial example is
found.

Deep Learning Verification (DLV)~\cite{HuangKWW17} is an approach that defines a
region of safety around a known input and applies SMT solving for checking robustness.
They consider the input space
to be discretized and alter the input using manipulations until it is
at a minimal distance from the original, to generate possibly-adversarial inputs. They can only guarantee freedom of adversarial perturbations within the discrete points that are explored.
Our clustering approach can potentially improve the technique by constraining the discrete search within regions.

   \section{Conclusion}
\label{sec:conclusion}

This paper presents a novel, data-guided technique to search for adversarial perturbations (or prove they cannot occur) within well-defined geometric regions in the input space that correspond to clusters of similar inputs known to share the same label. In doing so, the approach identifies and provides proof for regions of safety in the input space within which the network is robust with respect to target labels. Preliminary experiments on the ACAS Xu and MNIST datasets highlight the potential of the approach in providing formal guarantees about the robustness of neural networks in a scalable manner.

In the future, we plan to investigate the following directions:
\begin{enumerate}
    \item{\textbf{Re-training:}} If there are a number of single-input clusters or clusters with low cardinality, it could indicate two cases:
\begin{inparaenum}[(i)]
\item if the density is low, there is not enough training data in that region; or
\item if the density is high, there is a lot of noise or number of redundant attributes, which leads to repeated splitting of clusters.
\end{inparaenum}
 The first case could act as a feedback for re-training. The second
 case is an indicator that the clustering should probably be carried
 out at a higher layer of abstraction, with a smaller number of more
 relevant attributes. 
\item{\textbf{Input to other techniques:}} The boundaries of the
  cluster spheres formed by kMeans lie in low density
  areas. Therefore, the network could be assumed to have low
  accuracy around these boundaries. Thus, adversarial robustness
  checks around instances that are closer to the edge of the
  clusters could exhibit a high number of adversarial perturbations. This analysis
  could help identify potential inputs to which other techniques for
  assessing local robustness could be applied. 
\item{\textbf{Other Solvers:}} While in our implementation we have used Reluplex to perform the verification, our approach is general and can use other tools as a back-end solver. As checking robustness for deep neural networks is an active area of research, we plan to investigate and integrate other solvers, as they become available. We also plan to investigate testing, guided by the computed regions, as an alternative to verification, for increased scalability, but at the price of losing the formal guarantees.
\end{enumerate}

\medskip
\noindent
\textbf{Acknowledgements.}
This work was partially supported by
 grants from NASA, NSF, FAA and Intel.




\bibliographystyle{abbrv}
\bibliography{bib}

\begin{thebibliography}{10}

\bibitem{AggarwalHK01}
C.~C. Aggarwal, A.~Hinneburg, and D.~A. Keim.
\newblock On the surprising behavior of distance metrics in high dimensional
  spaces.
\newblock In {\em Proc. 8th Int. Conf. on Database Theory (ICDT)}, pages
  420--434, 2001.

\bibitem{CaKaBaDi17}
N.~Carlini, G.~Katz, C.~Barrett, and D.~Dill.
\newblock {Ground-Truth Adversarial Examples}, 2017.
\newblock Technical Report. \url{http://arxiv.org/abs/1709.10207}.

\bibitem{Carlini017}
N.~Carlini and D.~Wagner.
\newblock Towards evaluating the robustness of neural networks.
\newblock In {\em Proc. 38th IEEE Symposium on Security and Privacy}, 2017.

\bibitem{Ehlers2017}
R.~Ehlers.
\newblock Formal verification of piece-wise linear feed-forward neural
  networks.
\newblock In {\em Proc. 15th Int. Symp. on Automated Technology for
  Verification and Analysis (ATVA)}, 2017.

\bibitem{KurakinGB16a}
R.~Feinman, R.~R. Curtin, S.~Shintre, and A.~B. Gardner.
\newblock Adversarial machine learning at scale, 2016.
\newblock Technical Report. \url{http://arxiv.org/abs/1611.01236}.

\bibitem{FeinmanCSG17}
R.~Feinman, R.~R. Curtin, S.~Shintre, and A.~B. Gardner.
\newblock Detecting adversarial samples from artifacts, 2017.
\newblock Technical Report. \url{http://arxiv.org/abs/1703.00410}.

\bibitem{GoodfellowSS14}
I.~J. Goodfellow, J.~Shlens, and C.~Szegedy.
\newblock Explaining and harnessing adversarial examples, 2014.
\newblock Technical Report. \url{http://arxiv.org/abs/1412.6572}.

\bibitem{HuangKWW17}
X.~Huang, M.~Kwiatkowska, S.~Wang, and M.~Wu.
\newblock Safety verification of deep neural networks.
\newblock In {\em Proc. 29th Int. Conf. on Computer Aided Verification (CAV)},
  pages 3--29, 2017.

\bibitem{ACASXU}
K.~Julian, J.~Lopez, J.~Brush, M.~Owen, and M.~Kochenderfer.
\newblock Policy compression for aircraft collision avoidance systems.
\newblock In {\em Proc. 35th Digital Avionics System Conf. (DASC)}, pages
  1--10, 2016.

\bibitem{KanungoMNPSW02}
T.~Kanungo, D.~M. Mount, N.~S. Netanyahu, C.~D. Piatko, R.~Silverman, and A.~Y.
  Wu.
\newblock An efficient k-means clustering algorithm: Analysis and
  implementation.
\newblock {\em IEEE Transactions on Pattern Analysis and Machine Intelligence},
  24(7):881--892, 2002.

\bibitem{KaBaDiJuKo17Reluplex}
G.~Katz, C.~Barrett, D.~Dill, K.~Julian, and M.~Kochenderfer.
\newblock Reluplex: An efficient {SMT} solver for verifying deep neural
  networks.
\newblock In {\em Proc. 29th Int. Conf. on Computer Aided Verification (CAV)},
  pages 97--117, 2017.

\bibitem{KaBaDiJuKo17FVAV}
G.~Katz, C.~Barrett, D.~Dill, K.~Julian, and M.~Kochenderfer.
\newblock Towards proving the adversarial robustness of deep neural networks.
\newblock In {\em Proc. 1st Workshop on Formal Verification of Autonomous
  Vehicles (FVAV)}, pages 19--26, 2017.

\bibitem{KuGoBe16}
A.~Kurakin, I.~Goodfellow, and S.~Bengio.
\newblock {Adversarial Examples in the Physical World}, 2016.
\newblock Technical Report. \url{http://arxiv.org/abs/1607.02533}.

\bibitem{MNIST}
Y.~LeCun, C.~Cortes, and C.~J.~C. Burges.
\newblock The {MNIST} database of handwritten digits.
\newblock \url{http://yann.lecun.com/exdb/mnist/}.

\bibitem{Moosavi-Dezfooli16}
S.~Moosavi{-}Dezfooli, A.~Fawzi, and P.~Frossard.
\newblock Deepfool: A simple and accurate method to fool deep neural networks.
\newblock In {\em Proc. IEEE Conf. on Computer Vision and Pattern Recognition
  (CVPR)}, pages 2574--2582, 2016.

\bibitem{PapernotM16}
N.~Papernot and P.~D. McDaniel.
\newblock On the effectiveness of defensive distillation, 2016.
\newblock Technical Report. \url{http://arxiv.org/abs/1607.05113}.

\bibitem{PapernotMJFCS16}
N.~Papernot, P.~D. McDaniel, S.~Jha, M.~Fredrikson, Z.~B. Celik, and A.~Swami.
\newblock The limitations of deep learning in adversarial settings.
\newblock In {\em Proc. 1st IEEE European Symposium on Security and Privacy
  (EuroS\&P)}, pages 372--387, 2016.

\bibitem{PuTa10}
L.~Pulina and A.~Tacchella.
\newblock An abstraction-refinement approach to verification of artificial
  neural networks.
\newblock In {\em Proc. 22nd Int. Conf. on Computer Aided Verification (CAV)},
  pages 243--257, 2010.

\bibitem{PuTa12}
L.~Pulina and A.~Tacchella.
\newblock Challenging {SMT} solvers to verify neural networks.
\newblock {\em AI Communications}, 25(2):117--135, 2012.

\bibitem{SzegedyZSBEGF13}
C.~Szegedy, W.~Zaremba, I.~Sutskever, J.~Bruna, D.~Erhan, I.~Goodfellow, and
  R.~Fergus.
\newblock Intriguing properties of neural networks, 2013.
\newblock Technical Report. \url{http://arxiv.org/abs/1312.6199}.

\end{thebibliography}

\end{document}